\pdfoutput=1

\documentclass[11pt]{article}
\usepackage{float}
\usepackage{placeins}
\usepackage[]{ACL2023}
\usepackage{graphicx}
\usepackage{times}
\usepackage{subcaption}
\usepackage{latexsym}
\usepackage[T1]{fontenc}
\usepackage{multirow}
\usepackage[utf8]{inputenc}

\usepackage{microtype}
\usepackage{booktabs}
\usepackage{inconsolata}

\expandafter\def\expandafter\UrlBreaks\expandafter{\UrlBreaks
  \do\a\do\b\do\c\do\d\do\e\do\f\do\g\do\h\do\i\do\j%
  \do\k\do\l\do\m\do\n\do\o\do\p\do\q\do\r\do\s\do\t%
  \do\u\do\v\do\w\do\x\do\y\do\z\do\A\do\B\do\C\do\D%
  \do\E\do\F\do\G\do\H\do\I\do\J\do\K\do\L\do\M\do\N%
  \do\O\do\P\do\Q\do\R\do\S\do\T\do\U\do\V\do\W\do\X%
  \do\Y\do\Z}

\usepackage{soulutf8}
\usepackage[normalem]{ulem}
\usepackage{xcolor}

\usepackage{enumitem}
\setlist{nolistsep}
\usepackage{microtype}

\iffalse 
    \usepackage[final]{proofing}
  \renewcommand\hl[1]{{#1}}  
   {\draftnote{\red{#2}}}
   \newcommand\redHL[1]{}
  \newcommand\todo[1]{}

  \newcommand{\Djame}[1]{}

\else  

\usepackage[draft]{proofing}

\newcommand{\Djame}[1]{
\textbf{\textcolor{red}{\hl{Djame: #1}}}
}

\newcommand\red[1]{{{\textcolor{red}{\bf #1}}}}

\let\oldred\red
\renewcommand\red[1]{{ \oldred{{#1}}}}

 \newcommand\redHL[1]{\red{\hl{#1}}}
\let\olddraftnote\draftnote
\renewcommand\draftnote[1]{\olddraftnote{\red{#1}}}

\fi

\usepackage{microtype}
\usepackage{tabularx}
\usepackage{xspace}

\newcommand{\xlmt}{{\sc XLM-T}\xspace}
\newcommand{\xlmr}{{\sc XLM-R}\xspace}
\title{Cloaked Classifiers:  Pseudonymization Strategies on Sensitive Classification Tasks 

}

\author{Arij Riabi \quad Menel Mahamdi \quad Virginie Mouilleron \quad Djamé Seddah \\
     Inria, Paris\\
     \{firstname,lastname\}@inria.fr}

\begin{document}
\maketitle
\begin{abstract}

Protecting privacy is essential when sharing data, particularly in the case of an online radicalization dataset that may contain personal information. In this paper, we explore the balance between preserving data usefulness and ensuring robust privacy safeguards, since regulations like the European GDPR shape how personal information must be handled. We share our method for manually pseudonymizing a multilingual radicalization dataset, ensuring performance comparable to the original data. Furthermore, we highlight the importance of establishing comprehensive guidelines for processing sensitive NLP data by sharing our complete pseudonymization process, our guidelines, the challenges we encountered as well as the resulting dataset.

\end{abstract}

\section{Introduction}

Radicalization, fostered by online propaganda and offline indoctrination, has been the primary driver in most terror attacks and eruptions of public violence over the past decade \cite{farwell2014,fernandez2021artificial,PELLICANI2023435}. It can be defined as a process by which an individual or group adopts increasingly radical viewpoints in opposition to a political, social, or religious system \cite{fink2014understanding}. These viewpoints cover, for example, far-right ideologies, religiously inspired extremism, and extreme conspirationism. Such content can spread rapidly, especially through social media, making radicalization challenging to detect \citep{Nouh2019UnderstandingTR}. 

Natural Language Processing (NLP) methods have been used to detect and analyze radicalization mechanisms such as propaganda, recruitment, networking, data manipulation, and disinformation \cite{Torregrosa2021ASO,Aldera2021OnlineED,Gaikwad2021OnlineED}. However, the effectiveness of such detection models depends on the availability and quality of training and evaluation datasets. Protecting user privacy, especially for sensitive tasks, is imperative when sharing such datasets. Finding the right balance between the obligation to build accurate anonymization methods and the need to maintain a decent level of performance is hard,
as pertinent information may be contained through some identifiers (usernames, URLs, locations, etc.) and their associated socio-demographic or geographic markers. Hence, a {\em brutal} anonymization of a dataset can hinder its usability, especially in a domain where radicalization clues are often found through these indicators \cite{PELLICANI2023435}.   

 Ensuring the privacy of individuals is critical, especially in light of regulations such as the General Data Protection Regulation (GDPR)\footnote{The GDPR is a comprehensive data protection law enacted by the European Union (EU). It aims to protect the privacy and personal data of individuals within the EU and the European Economic Area (EEA). \iffalse GDPR regulates how organizations collect, store, process, and share personal data, giving individuals greater control over their personal information.\fi}. This is why we believe that despite implementing various laws to minimize harm and protect sensitive information, there is a need to explore how technological advancements intersect with data protection laws and impact the collection, storage, and use of confidential data \cite{10233989,lothritz-etal-2023-evaluating}. 


In this work, we present our methodology for the manual pseudonymization of a radicalization dataset that (i) ensures performance to be comparable to the original data while maintaining its semantic properties and (ii) protects user privacy.
We emphasize the importance of establishing a standard framework for privacy and usefulness when processing sensitive NLP data by sharing the complete pseudonymization process for our datasets and the challenges we faced \cite{vakili-dalianis-2022-utility,vakili-dalianis-2023-using}. It is a highly sensitive task that requires 100\% accuracy; any oversight can render the dataset invalid.

 Our dataset includes English, French, and Arabic content from various sources such as forums, Telegram and other social media platforms. The content covers different radicalization domains (from white supremacy to jihadism) for each language. Our dataset will be available upon publication\footnote{Note that evaluating the radicalization detection task in itself is not the main point of the paper; here, we focus on our pseudonymization process.}.

 
 The manual annotation process we devised guarantees a high level of precision and enables us to better explore the interaction of our NLP tools and improve user safety. Furthermore, a critical component of our methodology involves identifying the exceptions for which anonymization does not need to be applied. For example, keeping well-known events and public figures enables us to leverage the knowledge embedded in the language model about specific entities and prevent pseudonymization from corrupting the relationships and alignment between named entities and other elements within the text, thereby enhancing the effectiveness of our system. Our evaluation results show that models trained on our pseudonymized data maintain similar levels of performance to their original counterparts.
 
  To summarize, our contributions are as follows:
\begin{itemize}
    \item We developed and share detailed guidelines\footnote{\url{https://file.io/rmUwdPfvnmXq}} for our pseudonymization method.
    \item We release a pseudonymized multilingual radicalization detection dataset \footnote{\url{https://gitlab.inria.fr/ariabi/counter-dataset-public}}.
    \item We provide an analysis of performance, demonstrating that our method maintains the same level of effectiveness as the original data while protecting user privacy.
\end{itemize}

\section{Related Work}
\subsection{Definitions}
The GDPR provides a comprehensive definition of personal data, including any information related to an identified or identifiable natural person. According to Article 4 (1) of the GDPR, \textit{“personal data means any information relating to an identified or identifiable natural person (data subject); an identifiable natural person is one who can be identified, directly or indirectly, in particular by reference to an identifier such as a name, an identification number, location data, an online identifier or to one or more factors specific to the physical, physiological, genetic, mental, economic, cultural or social identity of that natural person”.}
Building on this definition, \textbf{anonymization} refers to the complete and irreversible removal of any data in a dataset that could potentially identify an individual, directly or indirectly. \textbf{De-identification} involves the removal of specific, predetermined direct identifiers from a dataset. \textbf{Pseudonymization} is replacing direct identifiers with pseudonyms or coded values while keeping the mapping between the pseudonyms and original identifiers stored separately. The definitions of these terms may vary across literature, and they are often used interchangeably \cite{lison-etal-2021-anonymisation,lothritz-etal-2023-evaluating}. 

Traditional manual methods for anonymizing text data may be inefficient, error-prone, and expensive, making it necessary to develop well-defined frameworks. \citet{lison-etal-2021-anonymisation}  point out a significant gap between NLP and privacy-preserving data publishing (PPDP) approaches, both of which have addressed aspects of anonymization independently without sufficient interaction \cite{papadopoulou-etal-2022-neural}. Given the complexity of text data, including indirect identifiers and nuanced semantic cues, there is a need for improved anonymization models that can effectively balance the trade-off between privacy protection and data utility. 

The NLP-based approach usually turns text anonymization into a NER-like problem \cite{eder-etal-2022-beste}, where a set of categories set in advance are to be retrieved from the text. The PPDP approach uses “privacy models” \cite{sanchez-david-2016,SANCHEZ201723,brown2022does}, which are sets of requirements that are to be met by the anonymization system, often regarding identification by aggregation of data, degrees of anonymization and potential attacks. 

\citet{yermilov-etal-2023-privacy} compare three machine-learning-based pseudonymization techniques that consist of a \textit{NER-based }classical approach, \textit{seq2seq} \cite{lewis-etal-2020-bart}, which frames the task as a sequence-to-sequence transformation using an encoder-decoder model, and \textit{LLM Pseudonymization}, which uses a two-step process with GPT-3 and ChatGPT: GPT-3 extracts named entities, and ChatGPT then pseudonymizes them.

Text pseudonymization usually requires three steps: (1) establishing relevant categories of personal data, (2) retrieving them, and (3) replacing them. We will briefly introduce the related works in the next subsections.

\subsection{Establishing categories}

To our knowledge, there is no standardized set of categories, especially for non-medical, unstructured, online textual data that is processed in the European Union.

Since pseudonymization has mainly been used in the medical domain, most papers use the Personal Health Identifiers (PHI) enumerated in the American HIPAA regulations \cite{HIPAA}, either as a reference \cite{YANG2015S30,dernoncourt2017identification} or as a starting point for further adaptation to the corpus \cite{VELUPILLAI2009e19,dalianis-velupillai-2010-certain,megyesi2018learner,eder-etal-2020-code}. 
Some draw categories from data observation \cite{medlock2006introduction,adams-etal-2019-anonymate,cetinoglu-schweitzer-2022-anonymising}. \citet{adams-etal-2019-anonymate} set 3 types of entities for their online chat corpus: Personal Identifying Information (PII), Corporate Identifying Information (CII), and Others, with only PII and CII being anonymized. 
Others create categories using the GDPR-based distinction between direct identifiers, indirect/quasi-identifiers, and sensitive data \cite{pilan-etal-2022-text,volodina-etal-2020-towards}.

Still, making up an all-encompassing set of categories is not an easy task, and when it comes to non-clinical data, the line between what is to be anonymized and what is not becomes blurred for some entities. \citet{cetinoglu-schweitzer-2022-anonymising} resorted to heuristics and highlighted the subjective dimension of data pseudonymization. 
The datasets often display some special categories that have to be mentioned and taken into account in the annotation scheme:
\begin{itemize}
    \item  Indirect or quasi-identifiers: they are almost always anonymized \cite{adams-etal-2019-anonymate,volodina-etal-2020-towards,lison-etal-2021-anonymisation} following the GDPR. An argument cited by many is the study conducted by \citet{sweeney2000}, which showed that 87\% of the US population could be identified only by zip code, date of birth, and gender. Moreover, Identification by data aggregation and its prevention is a common theme in the literature.
   \item  Sensitive information, such as ethnicity, political views or sexuality, are either anonymized or at least detected and annotated for further processing \cite{volodina-etal-2020-towards}.
    \item  Public figures: briefly mentioned in \citet{adams-etal-2019-anonymate} and \citet{cetinoglu-schweitzer-2022-anonymising}, they are not anonymized.
     \item Deceased people:  there has been no mention of the case of deceased people. Although GDPR doesn’t apply in this case, the French CNIL\footnote{French data protection authority.} has advised to apply data protection rules when it might impact families and close ones.
\end{itemize}

Finally, some have argued that one must not entirely rely on a closed, predefined set of categories: \citet{pilan-etal-2022-text} suggest that all textual elements must be considered, as they can still be used for re-identification, either directly or indirectly through inference.

\subsection{Data retrieval}

Data retrieval can be done manually or with rule-based models \cite{neamatullah2008automated,cetinoglu-schweitzer-2022-anonymising}, but most of the related works employ machine learning and, more recently, focus primarily on deep learning approaches \cite{dernoncourt2017identification,liu2017identification,papadopoulou-etal-2022-neural}.  Finally, anonymization pipelines and toolkits have also been proposed to coordinate human annotation and different anonymization techniques \cite{adams-etal-2019-anonymate,clos-etal-2022-pripa}.

\subsection{Substitution strategies}

Textual data substitution usually falls into three categories. 
One can choose categorization (a term first used by \citet{medlock2006introduction}), by which one exact string replaces all units from the same category. For example, the SOLID Twitter dataset \cite{rosenthal-etal-2021-solid} replaces all usernames with the placeholder “@USER,” and in \citet{volodina-etal-2020-towards}, all bank accounts are replaced by the same standardized string “0000-00 000 00”.
Another method we call non-realistic pseudonymization consists of replacing each unit with a specific identifier that does not mimic natural language. Such is the case in the Dortmund Chat Corpus 2.1 \cite{lungen_2017_1041873}, in which a person’s name is replaced by an id, such as \texttt{“[\_PERSONNAME-1\_]”}.
A third method, which we call realistic pseudonymization, attempts to avoid loss of linguistic information by replacing the unit with a semantically similar identifier and that mimics natural language \cite{cetinoglu-schweitzer-2022-anonymising,eder-etal-2022-beste,olstad-etal-2023-generation}. To preserve data quality, we chose this approach for our dataset.

Some research purpose to extend pseudonymization efforts beyond the clinical domain \cite{Lampoltshammer-2019,pilan-etal-2022-text,yermilov-etal-2023-privacy}. Nevertheless, these efforts are currently confined to a limited list of categories, such as names \cite{lothritz-etal-2023-evaluating} or just names and addresses \cite{accorsi2012}, in an artificial setting. We disclose the exhaustive list of entity categories and all the considerations taken into account during the anonymization for our task. 
Our position aligns with the recent research of \citet{szawerna-etal-2024-pseudonymization-categories}, who propose implementing a universal tagging system for categorizing personally identifiable information (PII) to improve pseudonymization processes. They emphasize that existing tagsets do not encompass all PII types found across various domains with the necessary level of detail for successful pseudonymization. 

The pseudonymization of our dataset is important for sharing it for research purposes, as it minimizes information loss, which is a well-known undesirable side effect  \cite{meystre2014can,sawhney2022user,lothritz-etal-2023-evaluating}. Additionally, \citet{Lampoltshammer-2019} showed that even small changes in data anonymization can significantly impact sentiment analysis results even though \citet{vakili-etal-2022-downstream} showed no significant change in performance after anonymization for clinical data. The results of our experiments that show almost no impact (Subsection \ref{subsec:results}) confirm their findings.

\section{Methodology}

We argue that the sensitive nature of certain tasks requires human annotators; therefore, a considerable amount of our pseudonymization process is done manually. Our guidelines are based on three primary sources: legal texts and recommendations from the French CNIL and the GDPR, existing research on data anonymization for NLP, and a thorough analysis of our corpus. As far as we know, no work has been published on the pseudonymization of radicalization data. We have also not found any official, standardized method for pseudonymizing textual data, neither from the GDPR/CNIL nor the literature.

\subsection{Data types}
 We define three main types of data in our dataset: data related to individuals, data related to organizations, and data related to content sharing.
 \paragraph{Data related to Individuals.}
 We have systematically anonymized all direct identifiers (e.g. names, addresses, email addresses, phone numbers) associated with private individuals. For indirect identifiers (e.g., nationality, general location, age, gender), we decided to anonymize at least one in cases where multiple identifiers appear in the same text.
 
  Following \citet{adams-etal-2019-anonymate,cetinoglu-schweitzer-2022-anonymising}, public figures are not anonymized. We also include journalists, politicians, and authors in that category. Additionally, we introduced a category for \textbf{``Influencers,''} determined by criteria such as social media presence, follower count, and appearances in mainstream media. Although these profiles are not anonymized, specific sensitive direct and indirect identifiers (e.g., personal phone numbers and addresses) are anonymized to ensure their safety.
 
 We balanced GDPR guidelines and CNIL advice for deceased individuals by not anonymizing deceased public figures while anonymizing private victims, in order to respect their memory and privacy. Regarding convicted individuals and terrorists, we excluded well-known and deceased terrorists from anonymization, considered age at the time of the crime, and anonymized those not found guilty or who underwent legal name changes, especially if they were minors.
 \paragraph{Data related to organizations.}
 We have chosen not to anonymize the names of organizations as a general practice. However, exceptions were made when the organization's name could serve as an indirect identifier of individuals, particularly those belonging to vulnerable groups or who might be targeted for their opinions. These cases include family/small businesses, companies providing specific religious services, student organizations based on ethnicity or religion, and workplaces of activists. Additionally, names of radical organizations displayed as usernames or group/channel names on social media were anonymized while preserving relevant semantic information. For instance, ``@ProudBoys-Massachusetts-admin'' (fictional) was transformed to "@Proud\_Boys\_MA\_main".

 \paragraph{Data related to content sharing.
}
In the dataset, content is typically shared through URLs and titles of media. When the content is considered too radical or too private to share, it is anonymized or invalidated as appropriate. This includes URLs redirecting to fundraising campaigns, personal blogs or websites of private individuals (e.g., Tumblr, WordPress), social media channels of radical groups (e.g., Telegram, Gab) along with their usernames, and URLs and titles of videos, movies, and songs produced by members of radical groups.


\subsection{Pseudonymization Pipeline}
\paragraph{Retrieval.} The first step was to use a fine-tuned model to generate NER pre-annotations automatically. This initial version of named entity annotations helped to extract aliases, individuals, and organizations. The model was fine-tuned on ANERcorp \cite{anercop2007,obeid-etal-2020-camel} for Arabic, FTB NER \cite{ortiz-suarez-etal-2020-establishing} for French, and CONLL2003 \cite{tjongkimsang2003conll} for English. Moreover, regular expressions were used to extract data that followed stable patterns, such as links, hashtags, and emails (Figure \ref{fig:entities_distribution} in Appendix \ref{subsec:dataset} for the distribution of the categories). Simultaneously, we fixed the silver NER annotations to add another layer of NER with a large tagset (See Table \ref{tab:entities}in Appendix \ref{subsec:dataset}).

\paragraph{Manual anonymization.} One annotator per language manually anonymized the entities and corrected pre-annotations. After each decision of anonymization was made, it was added to a token-level correspondence table for the languages to ensure that an entity has the same replacement across languages.
To maintain the cultural and stylistic integrity of the content while avoiding the disclosure of sensitive information, we attempted to choose pseudonyms mimicking the original names or aliases. This involved picking pseudonyms that shared a phonetic resemblance, incorporated special characters or numbers, considered linguistic nuances, included wordplay, maintained similar token length, or even incorporated details about the author's origins, perceived ethnicity and cultural references (see Table \ref{tab:example_replacement} in Appendix \ref{subsec:dataset}).

In some special cases where anonymization is not needed, such as for links and some specific usernames,  we use invalidation by adding changing characters. Re-identification can still be possible in these cases, but direct access is not.

Finally, we choose anonymization out of caution when in doubt\footnote{We did not calculate the inter-annotator agreement for the anonymization process, but we frequently discussed difficult decisions to ensure consistency. For NER, we calculated inter-annotator agreement with 100 randomly selected sentences in both English and French. The English annotator annotated 100 French sentences, and vice versa. The Cohen's Kappa Score for French was 0.9124 and for English was 0.8266, indicating a high level of agreement between annotators, suggesting closely aligned decisions.}.
\paragraph{Accounting for re-identification}
We carefully considered re-identification concerns, basing our anonymization efforts on established insights. Recognizing re-identification as a significant concern in PPDP, we accounted for the ``disclosure risk'' by considering the ``background knowledge'' a potential attacker might have, as described by \citet{sanchez-david-2016,SANCHEZ201723}. This background knowledge includes all web pages accessible through search engines. Consequently, our anonymization process considered all data types that could be used with search engines to identify an individual.

\section{Experiments}
In this section, we analyze the variation of the performance of the model in different scenarios and compare the use of anonymized data to original data for radicalization detection task.
\subsection{Tasks}
\paragraph{Radicalization Detection Task}
Our dataset includes English, French, and Arabic examples from various sources (Figure \ref{fig:dataset_stat} in Appendix \ref{subsec:dataset}), each with distinct characteristics. The English dataset contains messages from platforms like Telegram and forums, where radical groups promote their movements. The French dataset consists mainly of comments from social media platforms such as Twitter and Instagram, while the Arabic dataset primarily comprises religious texts focused on jihadism from sources like Facebook and Twitter. Those texts included a lot of deceased persons that were not anonymized. We had a different annotator for each language.

For our experiments, we focus on the annotation of \textit{Call for Action Classification} for English and French as their sizes are comparable, which entails categorizing content into one of five predefined levels based on the degree to which it motivates specific actions, ranging from ``negative'' to ``very high'' (See Appendix \ref{subsec:dataset} for more details).

\begin{table}[htb!]
\footnotesize
\centering
\begin{tabular}{@{}cccc@{}}\toprule
 &   Train& Dev&Test\\
  \cmidrule{2-4}
 & \multicolumn{3}{c}{\textit{English}}\\
\# examples &1735 & 194 & 484\\
\# anonymized entities& 1143&146 &326\\
\midrule
 & \multicolumn{3}{c}{\textit{French}}\\
\# examples  & 1888& 210 & 526 \\
\# anonymized entities&485 &51 & 158\\
\midrule
 & \multicolumn{3}{c}{\textit{Arabic}}\\
\# examples  & - & - & 1500 \\
\# anonymized entities& - & - & 130 \\
\bottomrule
\end{tabular}

\caption{ Statistics for English, French and Arabic}

\label{tab:data_stats}
\end{table}

\begin{table*}
\centering
\footnotesize
\begin{tabularx}{\linewidth}{@{}cX@{}}
\toprule
Original & Hit me up \textcolor{magenta}{@marie.delattre1}, \textcolor{cyan}{@handsomephilantropist} on Insta. Shoutout to \textcolor{olive}{Moshe Chaya}! At \textcolor{orange}{Rue Alphonse Metayer}.\\
\midrule
S0 & Hit me up, on Insta. Shoutout to ! At.! \\ 
S1& Hit me up \textcolor{magenta}{placeholder}, \textcolor{cyan}{placeholder} on Insta. Shoutout to \textcolor{olive}{placeholder}! At \textcolor{orange}{placeholder}. \\ 
S2 & Hit me up \textcolor{magenta}{username}, \textcolor{cyan}{username} on Insta. Shoutout to \textcolor{olive}{name}! At \textcolor{orange}{location}.\\ 
S3 & Hit me up \textcolor{magenta}{username11}, \textcolor{cyan}{usernsme22} on Insta. Shoutout to \textcolor{olive}{name44}! At \textcolor{orange}{location55}.\\ 
\midrule
Ours & Hit me up \textcolor{magenta}{@jane.doe1}, \textcolor{cyan}{@attractivehumanitariant} on Insta. Shoutout to \textcolor{olive}{Raj Avrom}! At \textcolor{orange}{Rue Hubert Couturier}.\\
\bottomrule
\end{tabularx}
\caption{Examples (Fictional) of different substitutions methods}
\label{tab:anonymization_substitutions}
\end{table*}
\subsection{Substitutions methods}
In this section, we evaluate our pseudonymization technique by comparing it to four methods from the existing literature \cite{10.1136/amiajnl-2012-001012,berg-etal-2020-impact}. We use metadata from our annotations to generate three additional anonymized dataset versions. The strategies we considered are as follows:
\begin{itemize}
    \item \textbf{Entity Deletion (S0)}
This method involves deleting the entity to anonymize it. While this approach maximizes privacy, it sacrifices data utility and coherence.

\item \textbf{Uniform Placeholder (S1)}
This method replaces all entities in the dataset with the same placeholder. It retains some data utility while ensuring anonymity but lacks category-specific differentiation.

\item \textbf{Category-Specific Placeholder (S2)}
Each category of entities (e.g., names, organizations) is replaced with a unique placeholder specific to that category across the dataset. This strikes a balance between anonymization and preserving some context-specific information.

\item \textbf{Unique Placeholder per Entity (S3)}
A unique placeholder is assigned to each entity in each document, maintaining sentence coherence while ensuring anonymity.
\end{itemize}
Table \ref{tab:anonymization_substitutions} shows the differences between the different automatic methods and our methods.

\subsection{Model training}
We fine-tune \xlmt \cite{barbieri-etal-2022-xlm}, an \xlmr \cite{conneau-etal-2020-unsupervised} model that has been fine-tuned on 200 million tweets (1\,724 million tokens) scraped between May 2018 and March 2020, in more than 30 languages. This model has been shown to be more adapted for social media data \cite{montariol-etal-2022-multilingual}. To ensure the reliability of our findings, we fine-tuned the model using five different seeds and reported the average performance across these five runs.

\begin{table*}[htb!]
\parbox{.55\linewidth}{
\centering
\footnotesize
\begin{tabular}{@{}cccc@{}}\toprule
\textbf{Training data} & Lang& Corresponding Test & Original Test\\\midrule
 Original &\multirow{6}{*}{en} &- & 64.63 \scriptsize{($\pm$2.0)}\\
  S0 & & 62.11\scriptsize{($\pm$3.5)} & 60.81\scriptsize{($\pm$3.3)} \\
 S1 & & 64.99\scriptsize{($\pm$1.5)} & 63.81\scriptsize{($\pm$1.1)}  \\
 S2 & &62.34\scriptsize{($\pm$2.6)} & 59.91\scriptsize{($\pm$2.8)} \\
 S3 && 65.55\scriptsize{($\pm$1.6)} & 63.50\scriptsize{($\pm$1.4)}  \\
 \midrule
 Ours & & 65.46\scriptsize{($\pm$1.0)} & 64.80\scriptsize{($\pm$2.2)} \\

\midrule
Original
  &\multirow{6}{*}{fr} & - & 65.65\scriptsize{($\pm$1.8)}  \\
S0 & & 64.13\scriptsize{($\pm$6.1)} & 66.78\scriptsize{($\pm$7.8)}  \\
 S1& & 65.89\scriptsize{($\pm$4.1)} & 66.41\scriptsize{($\pm$5.4)}  \\
S2 & & 63.52\scriptsize{($\pm$5.0)} & 62.31\scriptsize{($\pm$4.9)} \\
S3 & & 64.87\scriptsize{($\pm$4.2)} & 66.10\scriptsize{($\pm$4.5)}\\
\midrule
Ours & & 64.72\scriptsize{($\pm$4.8)} & 63.97\scriptsize{($\pm$4.3)} \\
\bottomrule
\end{tabular}

\caption{Results for each fine-tuned model on the original training and the different anonymized training sets when\textbf{ tested on the original test set (right)} and \textbf{the corresponding anonymized test sets (left)}. (Average Macro-F1 Scores over 5 Seeds) }

\label{tab:main_results}
}
\hfill
\parbox{.4\linewidth}{
\centering
\footnotesize
\begin{tabular}{@{}ccc@{}}\toprule
\textbf{Testing data} & Lang& Macro-f1\\\midrule
 Original & \multirow{6}{*}{en} &64.63\scriptsize{($\pm$2.0)}\\
  S0 & & 62.93\scriptsize{($\pm$2.0)}\\
 S1 & & 62.56\scriptsize{($\pm$2.1)}\\
 S2 & & 63.41\scriptsize{($\pm$2.6)}\\
 S3 & &  63.14\scriptsize{($\pm$1.9)}\\
 \midrule
 Ours & &  65.24\scriptsize{($\pm$2.7)}\\

\midrule
Original
  & \multirow{6}{*}{fr}&  65.65\scriptsize{($\pm$1.8)} \\
S0 & &  65.57\scriptsize{($\pm$3.5)}\\
 S1& &  65.46\scriptsize{($\pm$3.8)}\\
S2 & & 65.69\scriptsize{($\pm$3.6)}\\
S3 & & 65.86\scriptsize{($\pm$3.5)}\\
\midrule
Ours & & 67.88\scriptsize{($\pm$2.3)}\\
\bottomrule
\end{tabular}

\caption{Results for the model \textbf{trained on original data} and \textbf{tested on the test sets corresponding to different substitution methods} (Average Macro-F1 Scores over 5 Seeds)}

\label{tab:original_model_results}

}
\end{table*}
\subsection{Results}
\label{subsec:results}
For each language, we trained six models: four models for the automatically anonymized versions, one on the original data, and one on our anonymized version. 



Table \ref{tab:main_results} reports the average macro-F1 scores over 5 seeds for each fine-tuned model, evaluated on both the corresponding pseudonymized and original test sets.
Our approach resulted in a macro-F1 score of 65.46 for the English language models on the corresponding test set, which closely aligns with the highest score of 65.55 achieved by S3. This demonstrates the effectiveness of our method in maintaining data usefulness while ensuring robust anonymization. When evaluated on the original test set, our method achieved a score of 64.80, outperforming all other methods and slightly outperforming the model trained on the original data (64.63). This indicates that our method introduces minimal noise, thereby preserving data quality and coherence.

The performance of our pseudonymization technique shows different tendencies in the English and French language models. While our method performed consistently well for the English models, this trend was not observed for the French models. Our method demonstrated a good balance between anonymization and data utility for the French dataset. However, it did not consistently outperform other methods across the corresponding pseudonymized and original test sets. 

The differences in trends observed between the French and English datasets can be attributed to the unique content and characteristics of the data for each language. The English dataset primarily consists of messages from platforms like Telegram and forums such as 4chan, where radical groups actively promote their movements and share propaganda. The figures (Figure \ref{fig:dataset_stat} in Appendix \ref{subsec:dataset}) further illustrate these differences, showing the diverse range of platforms for the English dataset and a higher proportion of radical content compared to the French dataset. As a result, it contains a significantly higher number of usernames and links that need to be anonymized. In contrast, the French dataset mainly includes posts from social media platforms like Twitter and Instagram. While personal data is less frequently encountered in the French dataset, it requires equal vigilance due to the presence of sensitive information, such as personal addresses and family business details. Table \ref{tab:data_stats} shows the distribution of the categories for both languages and total entities for the test sets.

\paragraph{What to use for training?} A commonly asked question after pseudonymization is, should we use the pseudonymized version for training? Does the added noise make the training more robust? Recent model attacks have demonstrated that it is possible to extract training data from a publicly shared model \cite{song-2017,carlini2021extracting}. To investigate this question, we report in Table \ref{tab:original_model_results} the results of models trained on the original training data and tested on each version of the pseudonymized test set similarly to \citet{lothritz-etal-2023-evaluating}. We do not observe the same tendencies for both languages. For English, training on the anonymized train set (Table \ref{tab:main_results}, corresponding test set column) gave better results than the counterpart model trained on the original data for almost half the models. While the results were inconsistent for English, we noticed that the original model performed consistently better in almost all cases when tested on the anonymized test sets for French. This suggests that the model learns more easily on the original data and generalizes well on the pseudonymized test sets.

Despite those trends, \citet{brown2022does} argue that language models should be trained on data that can be publicly published to guarantee privacy.

Even though it is not the main topic of this paper, we present in Table \ref{tab:main_ner} in Appendix \ref{subsec:additional_results} the results for the NER task on the original data and our anonymized data. We opted not to conduct experiments on the automatic substitution strategies because adding the category of the entity provides the named entity in the text, and removing it alters the token count, making the results non-comparable. We observe similar performance trends to the classification task with very close scores between the model trained on the original data and the model trained on our pseudonymized data.
\section{Challenges}

\paragraph{Public figures and influencers} The lines between public figures, ``influencers'', and ``private figures'' are often blurred, making it challenging to determine if a journalist for a small news website should be considered a public figure. Similarly, categorizing scholars and less renowned authors also poses difficulties.
\paragraph{Links redirecting towards radicalized content and far-right media websites}
It was often tough to decide what was to be anonymized for two reasons: the definition of “mainstream” can become entirely subjective, especially when a medium can be considered renowned in its circle but not enough for global recognition. Moreover, even when a medium is categorized as mainstream, leaving it as such still poses an ethical dilemma, as it can contribute to sharing propaganda. 
\paragraph{Data related to terrorists and attackers}

In the English and mainly Arabic datasets, there were a lot of names of deceased terrorists, mainly from the Far-Right or from ISIS. 
While it is common for ISIS terrorists to have acquired names that do not always correspond to their birth names, and thus the risk of identification is lower, it is still a dilemma as to what should be left in the dataset. 

\section{Conclusion}
In this paper, we presented our approach to pseudonymization specifically tailored for a radicalization dataset. Our method aimed to fill the gap in research on pseudonymization in sensitive domains, such as online radicalization. Our technique balances the need for privacy protection while maintaining the usefulness of the data for research and analysis. We highlighted the challenges encountered during the pseudonymization process, particularly the nuances of handling different types of personal data. These challenges underscore the importance of a detailed and cautious approach.  Our multilingual radicalization dataset will be released upon publication. We advocate for developing a standardized framework for pseudonymizing sensitive NLP data. Overall, our work contributes to the growing body of research advocating for enhanced privacy measures in the processing and sharing of sensitive data, aligning with recent efforts to establish universal standards for categorizing and anonymizing personally identifiable information \cite{szawerna-etal-2024-pseudonymization-categories}.
\section*{Limitations}
\paragraph{Legal implications of pseudonymization}
Social media data processing and publishing cannot be exempt from anonymization techniques. Article 4 of GDPR defines pseudonymization as \textit{“the processing of personal data in such a manner that the personal data can no longer be attributed to a specific data subject without the use of additional information, [...]”}, which \textit{“ is kept separately and is subject to technical and organizational measures [...].”}. This “additional information” is often shaped through correspondence tables between the original data and its pseudonymized counterpart. Pseudonymization is recommended by GDPR (art.89) as an example of “appropriate safeguard[s]” to process personal data.
Pseudonymization is not a completely fireproof method. According to the CNIL (2022)  and GDPR, personal data can still be recovered by accessing the correspondence tables or tertiary data. Thus, since private information can theoretically be recovered, pseudonymized data still falls under GDPR.

\section*{Ethics Statement}

This paper aims to outline the challenges encountered during the pseudonymization of this dataset. We share the resultant dataset as a scientific artifact in line with the principles of open science. We cannot stress enough This dataset cannot be used to train any radicalization model used in real ground conditions. Having been annotated by domain experts from different countries, it may contain biases that can harm different communities.

We recognize the sensitive nature of this work and stress the importance of striking a balance between privacy and effectiveness. We understand that the task of detecting radicalization is inherently subjective. Although we chose not to anonymize information about public figures, we took special care to anonymize contact and address information to prevent doxxing. For example, in one case from the English dataset, an individual with a somewhat public status in academia had their personal information -such as professional email addresses and phone numbers- revealed by the author of the post to incite harassment due to the individual’s political beliefs. Despite the public status of the individual, we determined that it was too dangerous to keep this information in the dataset.


Note that the whole annotation process was particularly challenging for our annotators due to the violent, if not borderline traumatizing in some cases, nature of the data, which had an impact on their psychological well-being. 

A mental health professional service and support from human resources services were made available to the team. A process dedicated to evaluating the psychological impact induced by annotating this content was put in place. Its results (through extensive surveys—similar in depth to PTSD evaluation forms—and debriefing interviews) are currently under evaluation at our institution.   

 \section*{Acknowledgements}
This work received funding from the European Union’s Horizon 2020 research and innovation program under grant agreement No. 101021607. The authors warmly thank the OPAL infrastructure from Université Côte d'Azur for providing resources and support.

\bibliography{anthology,custom}
\bibliographystyle{acl_natbib}

\appendix

\section{Appendix}
\label{sec:appendix}

\subsection{Datasets}
\label{subsec:dataset}
Each document of the original dataset is annotated with different information. We describe here the  \textbf{Call for Action levels} that indicates whether a specific content should be flagged:

\begin{itemize}
    \item \textbf{Negative (No Call for Action)}: Content that exhibits no indications of radicalization or encouragement of extremist activities.
    \item \textbf{Low Call for Action}: Content that expresses radical views or ideologies without explicitly advocating for violence or extremist actions. This may include mere approval of extremist actions or actors.
    \item \textbf{Moderate Call for Action}: Typically involves content that subtly suggests participation in extremist activities or ideologies but stops short of direct advocacy.
    \item \textbf{High Call for Action}: Content that demonstrates clear support or admiration for extremist groups or indicates involvement in such groups’ activities, likely inciting further radical actions.
    \item \textbf{Very High Call for Action}: Represents the most extreme level, where content explicitly calls for violent action against individuals or groups.
\end{itemize}

\begin{figure*}[!ht]
    \centering
    \begin{subfigure}[b]{0.49\textwidth}
        \centering
        \includegraphics[width=\textwidth]{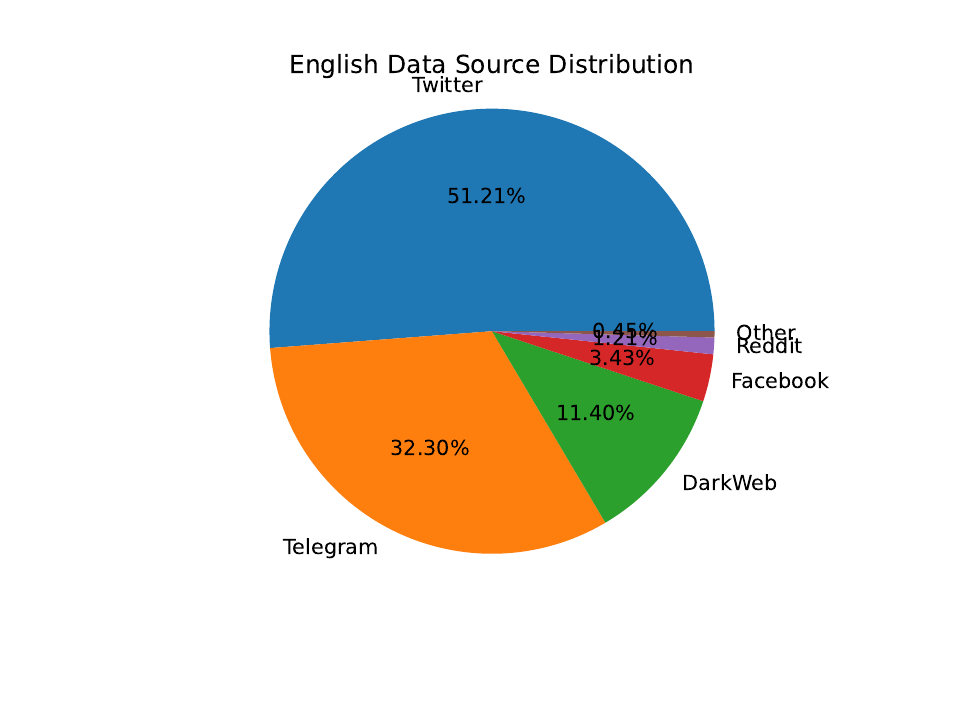}
        \caption{English Data Source Distribution}
    \end{subfigure}
    \begin{subfigure}[b]{0.49\textwidth}
        \centering
        \includegraphics[width=\textwidth]{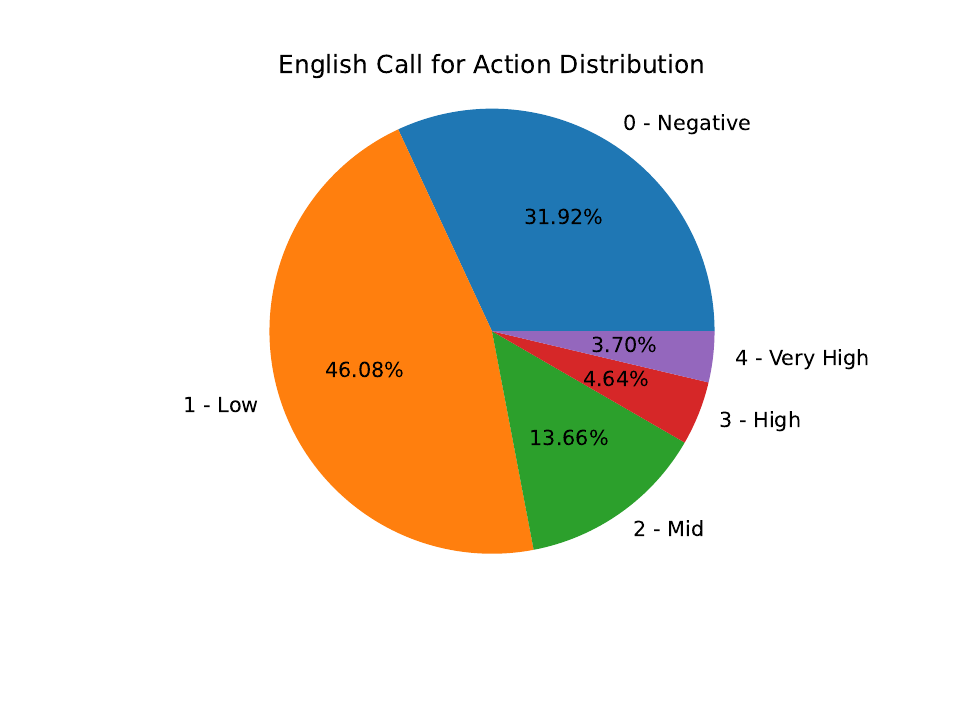}
        \caption{English Call for Action Distribution}
    \end{subfigure}

    \vspace{1em} %

    \begin{subfigure}[b]{0.49\textwidth}
        \centering
        \includegraphics[width=\textwidth]{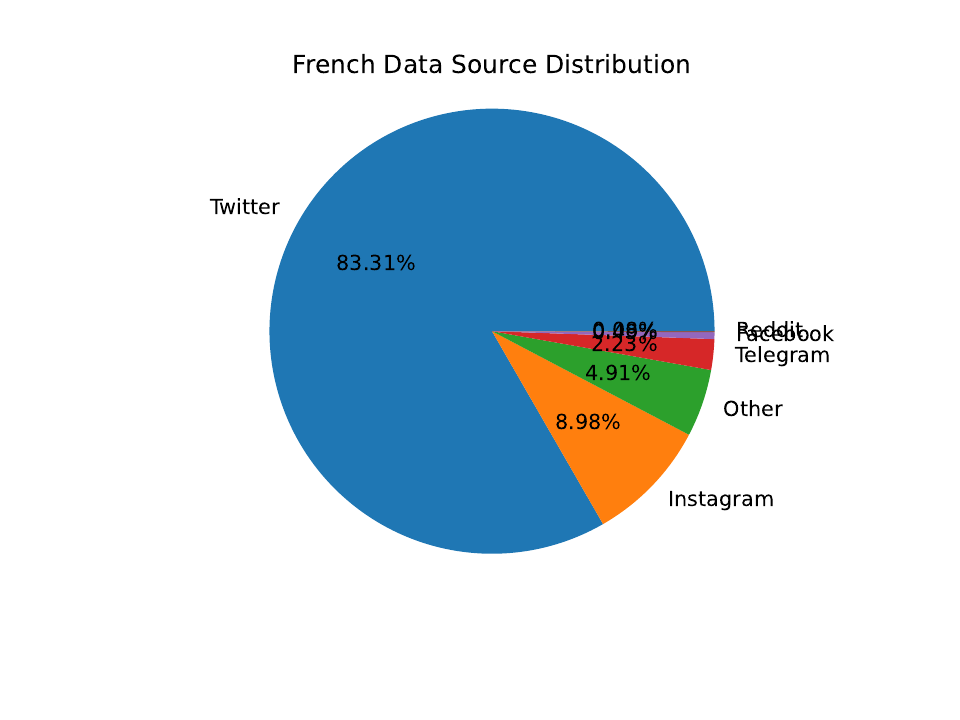}
        \caption{French Data Source Distribution}
    \end{subfigure}
    \hfill
    \begin{subfigure}[b]{0.49\textwidth}
        \centering
        \includegraphics[width=\textwidth]{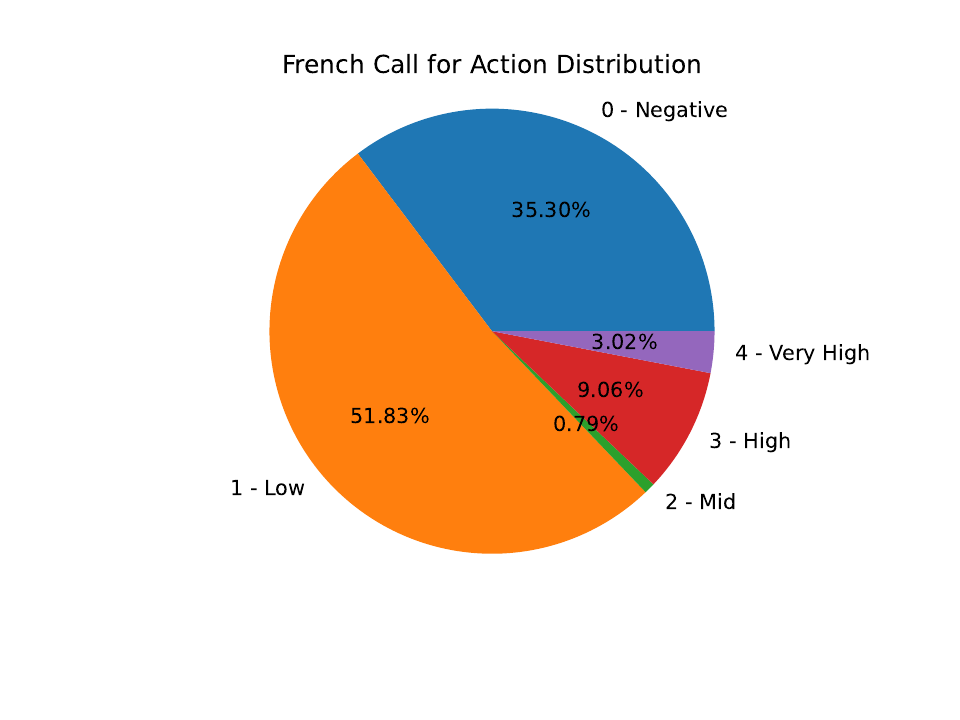}
        \caption{French Call for Action Distribution}
    \end{subfigure}

\vspace{1em} 

    \begin{subfigure}[b]{0.49\textwidth}
        \centering
        \includegraphics[width=\textwidth]{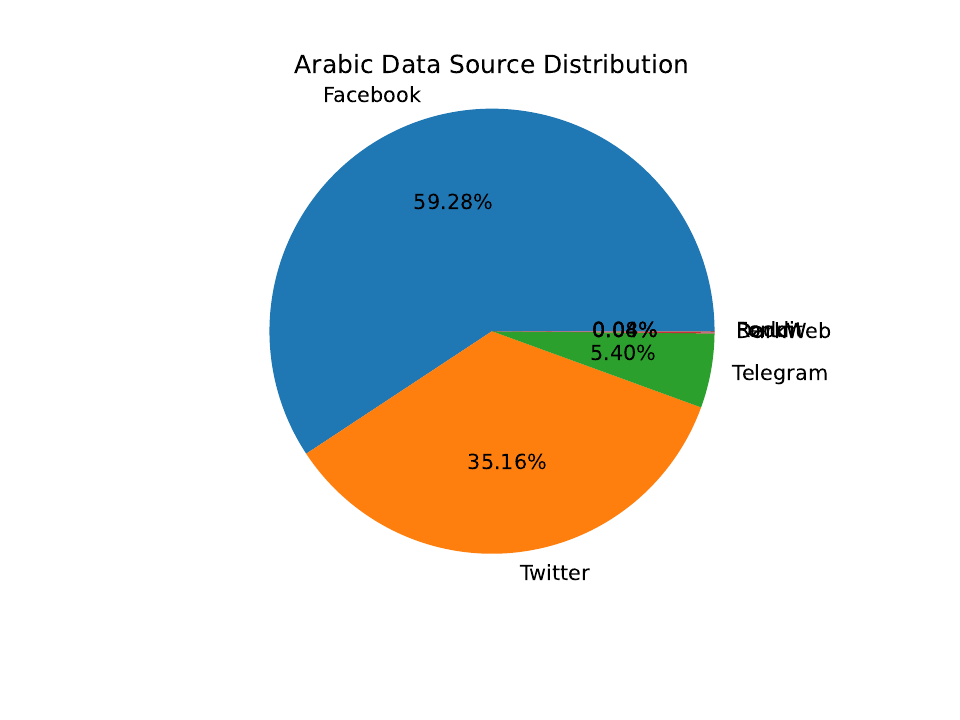}
        \caption{Arabic Data Source Distribution}
    \end{subfigure}
    \hfill
    \begin{subfigure}[b]{0.49\textwidth}
        \centering
        \includegraphics[width=\textwidth]{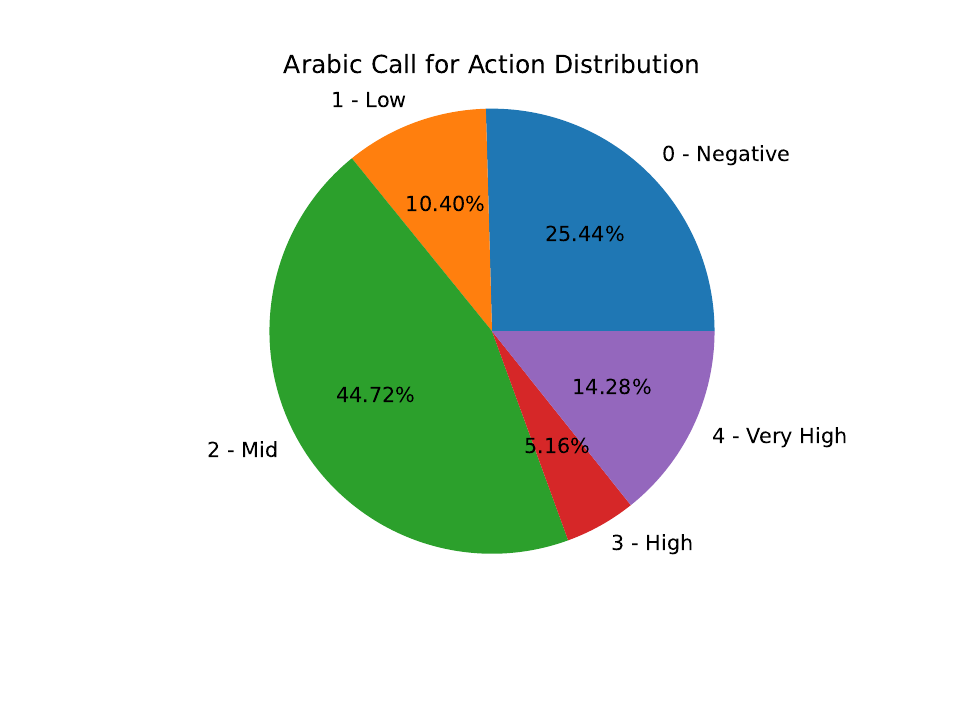}
        \caption{Arabic Call for Action Distribution (}
    \end{subfigure}
    \caption{Data source and call for action distributions for English, French, and Arabic}
    \label{fig:dataset_stat}
\end{figure*}

\begin{figure}[H]
    \centering
    \includegraphics[width=0.5\textwidth]{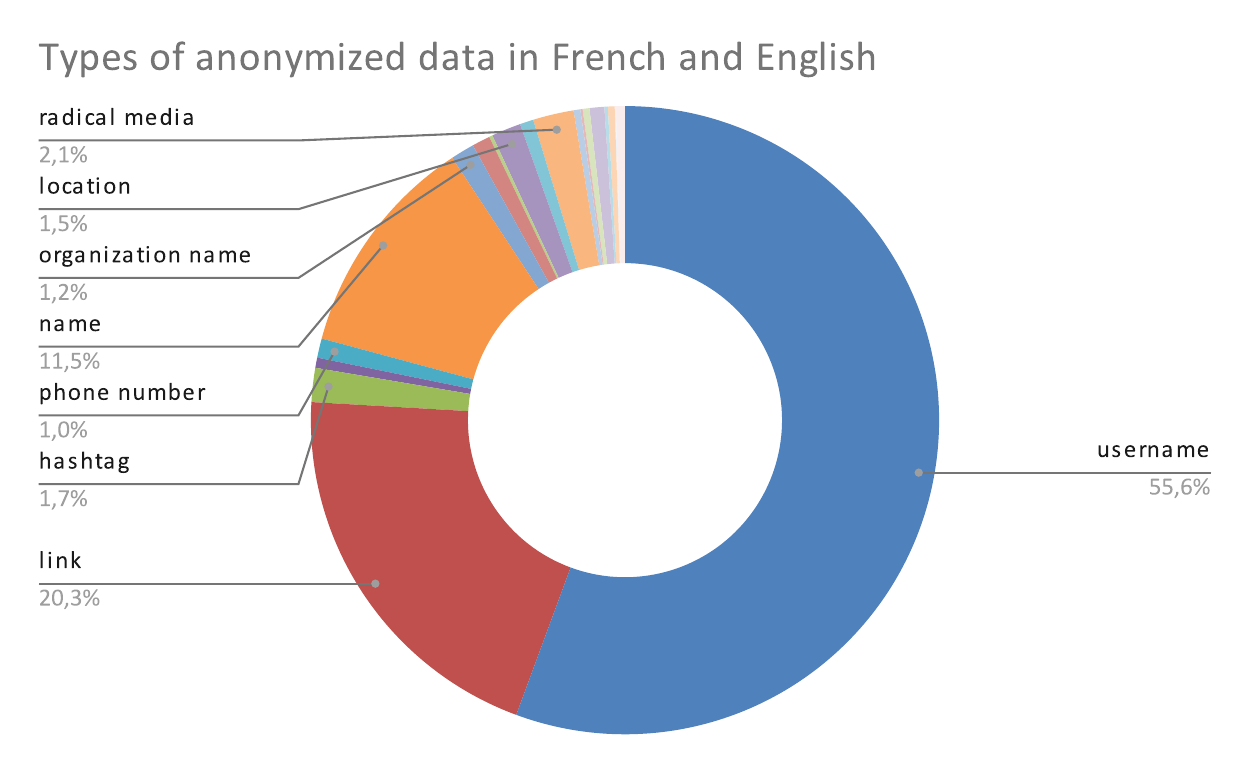}
    \caption{Types of anonymized data in French and English}
    \label{fig:entities_distribution}
\end{figure}

Figure \ref{fig:entities_distribution}, Figure \ref{fig:dataset_stat}, Table \ref{tab:ner_distribution}, Table \ref{tab:example_replacement} and Table \ref{tab:entities} represent statistics on our dataset and details about the annotations layers.

\begin{table}[h!]
\centering
\footnotesize
\begin{tabular}{lllll}
\toprule
 & English & French & Arabic \\
\midrule
PER & 2234 & 1802 & 4100 \\
LOC & 1783 & 1496 & 1656 \\
ORG & 1963 & 681 & 637 \\
OTH & 613 & 783 & 180 \\
COMP & 58 & 122 & 6 \\ 
\bottomrule
\end{tabular}
  \caption{Named entity repartition in the datasets.}
  \label{tab:ner_distribution}
\end{table}
\begin{table}[H]
    \centering
\footnotesize
    \begin{tabular}{cc}
    \toprule
       Original  &  Replacement \\
       \midrule
       Myriam Zegman  & Rachel Kaufman\\
       Virginia & Mary\\
       Muhammed & Ahmed\\
       @MaryJohanson1987& @LaraWilson1989\\
      https://wa.me/+93722758 &https://wa.me/+93824556\\
       \bottomrule
    \end{tabular}
    \caption{Examples (fictional) of replacements}
    \label{tab:example_replacement}
\end{table}

\begin{table*}[!ht]
\footnotesize
\centering
\begin{tabular}{lp{12cm}}
\toprule
\textbf{Label} & \textbf{Description} \\ \midrule
PER & mentions of names, aliases, and hashtags when they refer to a single person or user \\ 
PER:IMG & Fictional characters from manga, movies, books, and common culture. \\ 
PER:REL & References to individuals existing in a religious representation of the world. \\ 
COMP & Mentions of commercial enterprises and companies. \\ 
LOC & Mentions of locations, including neighborhoods, cities, and countries. \\
LOC:IMG & Fictional places. \\ 
LOC:REL & Religious locations.\\ 
ORG & Political, educational, or association-like organizations. \\ 
ORG:MEDIA & Media organizations, including radio or TV shows, podcasts, and newspapers. \\ 
OTH:BOOK & Books, mostly religious texts such as the Quran and the Bible. \\
OTH:GAME & References to games with mentions like "Minecraft." \\ 
OTH:MOVIE & Movies and series. \\ 
OTH:MUSIC & Musical entities, with mentions like "La isla Bonita." \\ 
OTH:DIS & Diseases. \\ 
OTH:SYMB & This category encompasses symbolic entities, including representations like the "Swastika" and religious symbols like the "Étoile de David." \\ 
OTH:EVENT & Reserved for recurring events, historical events, and religious events \\ 
OTH:CONSPI & This category is dedicated to concepts related to conspiracy theories.\\ 
\bottomrule
\end{tabular}
\caption{List of Named Entities used for the NER annotation layer.}
\label{tab:entities}
\end{table*}
\subsection{Additional Results}
\label{subsec:additional_results}

\begin{table*}[!ht]
\parbox{.55\linewidth}{
\centering
\footnotesize
\begin{tabular}{@{}cccc@{}}\toprule
Training data & Lang& Corresponding Test & Original Test\\\midrule
 Original &\multirow{2}{*}{en} & - & 87.04\scriptsize{($\pm$0.6)} \\
 Ours & & 87.01\scriptsize{($\pm$0.5)} &86.83\scriptsize{($\pm$0.5)} \\

\midrule
Original
  &\multirow{2}{*}{fr} & - & 78.96\scriptsize{($\pm$1.9)}  \\
Ours & & 78.96\scriptsize{($\pm$1)} & 78.01\scriptsize{($\pm$1.1)} \\
\bottomrule
\end{tabular}

\caption{NER results for each fine-tuned model on the original training and our anonymized training sets when\textbf{ tested on the original test set (right)} and \textbf{our anonymized test set (left)}. (Average Macro-F1 Scores over 5 Seeds) }

\label{tab:main_ner}
}
\hfill
\parbox{.4\linewidth}{
\centering
\footnotesize
\begin{tabular}{@{}ccc@{}}\toprule
Testing data & Lang& Macro-f1\\\midrule
 Original & \multirow{2}{*}{en} &87.04\scriptsize{($\pm$0.6)}\\
 Ours & &  86.01\scriptsize{($\pm$0.8)}\\

\midrule
Original
  & \multirow{2}{*}{fr}&  78.96\scriptsize{($\pm$1.9)} \\
Ours & & 77.87\scriptsize{($\pm$1.5)}\\
\bottomrule
\end{tabular}

\caption{NER results for the model \textbf{trained on original data} and \textbf{tested on our anonymized test set} (Average Macro-F1 Scores over 5 Seeds)}

\label{tab:original_model_ner}

}
\end{table*}
\end{document}